\documentclass{article}
\usepackage{iclr2027_conference,times}
\usepackage[T1]{fontenc}

\usepackage{amsmath}
\usepackage{amssymb}
\usepackage{amsfonts}
\usepackage{mathtools}
\usepackage{graphicx}
\usepackage{xcolor}
\usepackage{xspace}
\usepackage{booktabs}
\usepackage{multirow}
\usepackage{adjustbox}
\usepackage{hyperref}
\usepackage{etoolbox}
\makeatletter
\def\fg@nestbody#1{\vbox{\vbox{\vbox{\vbox{\vbox{\vbox{\vbox{\vbox{#1}}}}}}}}}
\patchcmd{\@outputpage}{\box\@outputbox}{\fg@nestbody{\box\@outputbox}}{}%
  {\PackageWarningNoLine{main}{Could not patch \string\@outputpage}}
\makeatother
\usepackage{url}
\usepackage[ruled]{algorithm2e}

\SetAlFnt{\small}
\SetAlCapFnt{\small}
\SetAlCapNameFnt{\small}
\SetAlCapHSkip{0pt}

\makeatletter
\DeclareRobustCommand\onedot{\futurelet\@let@token\@onedot}
\def\@onedot{\ifx\@let@token.\else.\null\fi\xspace}
\def\eg{e.g\onedot,\xspace}

\makeatother

\definecolor{darkgreen}{rgb}{0,0.7,0}
\definecolor{lightred}{rgb}{1.,0.5,0.5}

\newcommand{\myparagraph}[1]{\vspace{0.1cm}\noindent\textbf{#1}}
\newcommand{\mysuppparagraph}[1]{\noindent\textbf{#1}}
\newcommand{\moniker}{FluidGen\xspace}
\newcommand{\wo}{w\slash o\xspace}

\AtBeginDocument{\let\cite\citep}

\renewenvironment{figure*}[1][tbp]{\begin{figure}[#1]}{\end{figure}}
\renewenvironment{table*}[1][tbp]{\begin{table}[#1]}{\end{table}}
\newenvironment{teaserfigure}{\begin{figure}[!ht]}{\end{figure}}

\title{FluidGen: Physics-Aware Fluid Video Generation via Motion-Guided Post-Training}

\author{Yang Zheng$^{*}$ \quad Zizhang Li$^{*}$ \quad Boyang Deng$^{*}$ \quad Jan Ackermann \quad Lei Shu \\
\textbf{Yinghao Xu \quad Jiajun Wu \quad Gordon Wetzstein \quad Leonidas Guibas} \\
\textnormal{Stanford University} \\
\textnormal{$^{*}$Equal contribution}}

\iclrfinalcopy

\begin{document}

\maketitle
% \lhead{Preprint}

\begin{teaserfigure}
  \centering
  \includegraphics[width=\linewidth]{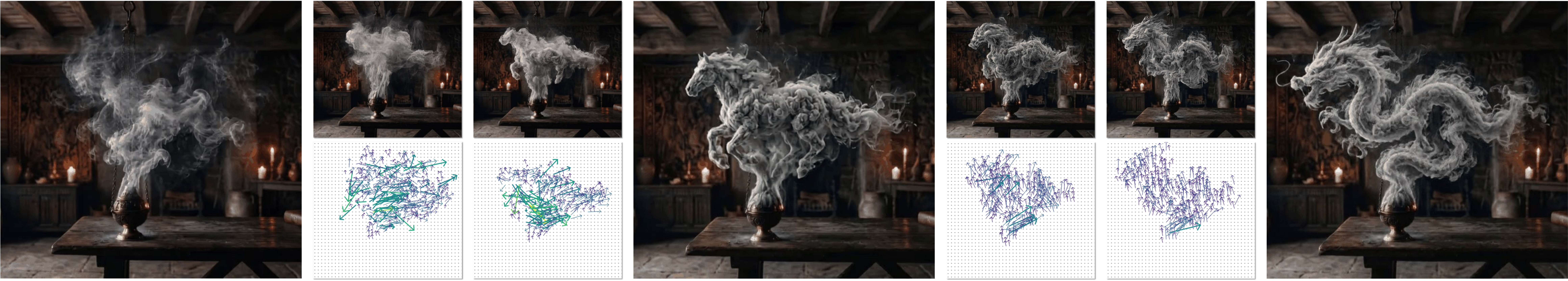}
  \caption{We introduce \textbf{FluidGen}, a physics-aware video diffusion model for realistic fluid dynamics. FluidGen jointly predicts per-frame velocity fields (insets) that regularize generation toward physically consistent motion.
    }
\label{fig:teaser}
\end{teaserfigure}

\begin{abstract}

Video generation models produce visually compelling motion but systematically fail on fluid phenomena, including smoke, fire, and explosion debris, where outputs lack coherent vortical structure and collapse to bland dynamics that violate the physics of real fluids. Classical solvers achieve high physical fidelity but require detailed 3D scenes and carefully tuned parameters, making them impractical for in-the-wild imagery and video generation. We address this gap by injecting a data-driven fluid prior into a pre-trained video model, trained on a mixture of high-fidelity Houdini simulation data with paired velocity-field ground truth and curated in-the-wild fluid video data for appearance diversity. The core of our approach is a dual stream architecture that predicts per-frame velocity fields alongside RGB, supervised by simulation ground truth. This auxiliary signal regularizes generation toward physically consistent motion and exposes an interpretable physical quantity that we further leverage as a reward for velocity-aware direct preference optimization. Our model produces substantially more realistic fluid dynamics sequences than strong video baselines and optimization-based methods. Our data, code, and model will be publicly released. Project page: \url{https://y-zheng18.github.io/FluidGen/}.

\end{abstract}

\section{Introduction}

Realistic fluid phenomena, including billowing smoke, rolling fire, and the dust and ash of an explosion, are among the most important visual elements in film, games, and immersive media, and also among the hardest to make look right. Their motion is driven by complex, turbulent physical dynamics, and humans can notice errors in problematic fluid dynamics immediately: for example a plume that rises too uniformly, a flame that moves in an unnatural direction, or a debris cloud without proper swirling structure. Synthesizing fluid content that holds up under this realism has been a long-standing goal in computer graphics, and despite decades of progress, it still typically requires expert artists, expensive simulation tools, and large amounts of manual effort.

One family of approaches treats fluid dynamics synthesis as a simulation problem. Classical physical solvers solve the Navier--Stokes equations~\cite{constantin1988navier} on grids~\cite{stam2023stable,bridson2015fluid}, particles~\cite{muller2003particle,zhu2005animating}, or hybrid representations~\cite{macklin2013position}, and produce physically faithful results when the scene parameters and boundary conditions are carefully defined. To extend simulation to single- or multi-view videos captured in the wild, more recent works reconstruct a 3D fluid representation from videos with regularization derived from physical priors \cite{eckert2019scalarflow,chu2022physics,yu2023inferring,gao2025fluidnexus} and re-simulating with the reconstructed representation. Another line of work uses neural networks as replacements to accelerate explicit solvers~\cite{kim2019deep,sanchez2020learning, ummenhofer2019lagrangian,cai2021physics}. A shared fundamental component of these works is the reliance on manually designed simulation rules~(\eg Navier--Stokes equations) that are themselves only approximations of the highly complex fluid dynamics process. As a result, they remain brittle: small reconstruction errors build up, fluid parameters must be tuned per scene, and the state-of-the-art methods still struggle on the casual images that dominate real-world content.

A new family of approaches turns to large-scale generative models, which now carry remarkably strong priors over both motion and appearance from internet-scale training~\cite{wan2025wan,sora,polyak2024movie,bar2024lumiere}. A natural way to use them in physical content creation is as a learned renderer on top of an explicit simulator, leaning on the generator's appearance prior while leaving the dynamics to the solver. Yet on fluid phenomena specifically, this combination still falls short. On their own, modern video models consistently violate basic physical rules in scenes with complex fluid dynamics~\cite{motamed2025generative}. Earlier physics-aware generation efforts have focused mostly on rigid~\cite{liu2024physgen} or elastic~\cite{tan2024physmotion} objects rather than on volumetric, turbulent fluids. To close this gap on fluid dynamics, very recent works \cite{li2025wonderplay,zhan2026perpetualwonder,liu2026realwonder} explicitly couple a video generator with an underlying simulator that provides fluid dynamics guidance. However, all aforementioned pipelines still depend on the heavy machinery of an authored 3D scene and a configured simulator, which limits their applicability to in-the-wild fluid content synthesis.

We argue that video model itself can learn complex fluid dynamics directly, but RGB supervision alone provides an insufficient training signal: appearance errors dominate the convergence while subtle physical errors are easily ignored. We therefore introduce an internal motion representation, jointly predicted with RGB, that explicitly exposes fluid dynamics to the model. Specifically, we predict a 3D velocity field describing fluid motion over time, providing a more physically meaningful representation than optical flow or trajectories~\cite{wang2024motionctrl,zhang2025tora}.

This representation enables a clean two-stage post-training pipeline for a pretrained video diffusion model~\cite{wan2025wan}. First, we jointly supervise RGB and velocity prediction using high-fidelity Houdini simulations with ground-truth velocity fields, together with in-the-wild fluid videos for appearance diversity and gives the model the generalization ability for in-the-wild content. Second, we perform velocity-aware DPO~\cite{rafailov2023direct,liu2025videodpo}, using the model's own predicted velocity field together with appearance signals to rank samples by physical plausibility. The resulting model produces more coherent vortices, natural advection, and fine-scale turbulent structure and clearly outperforms state-of-the-art strong video-generation and simulation-based baselines.

In summary, our contributions are:

\begin{itemize}
    \item We propose a dual-head video model architecture that jointly predicts an internal fluid velocity field alongside RGB pixels, providing an explicit and interpretable physical signal that guides dynamics learning.
    \item We curate a fluid training corpus that pairs high-quality Houdini simulations with collected in-the-wild smoke, fire, and explosion footage, combining dense physical supervision with the appearance variety needed for real-world generalization.
    \item We design a two-stage training pipeline that first jointly supervises RGB and velocity prediction and then aligns the model outputs with a post-training stage using its own predicted velocity field as the reward.
\end{itemize}

\section{Related Work}

\myparagraph{Fluid Simulation and Reconstruction.}
Realistic fluid visual effect synthesis in graphics has long been driven by analytical physical simulation. Classical Eulerian solvers discretize the Navier--Stokes equations on regular grids~\cite{stam2023stable,bridson2015fluid}, while particle-based formulations~\cite{muller2003particle,zhu2005animating} represent fluids as collections of moving particles, and hybrid representations such as PBF~\cite{macklin2013position} balance accuracy with interactivity from both sides. To reduce simulation cost, neural networks have been used as accelerating components for the traditional solvers~\cite{kim2019deep,sanchez2020learning,ummenhofer2019lagrangian,cai2021physics}. Built on top of existing fluid simulation formulations, a complementary line of work aims to recover fluid representations directly from captured video and re-simulates them~\cite{eckert2019scalarflow,chu2022physics,yu2023inferring,gao2025fluidnexus,li2025smokesvd,liu2025wildsmoke}. Combined with differentiable rendering, these works directly build fluid assets from real-world captures instead of requiring manual construction in the simulator. However all of these methods rely on human-defined simulation rules and are bounded by the approximation accuracy. Our work takes the opposite path: we treat fluid dynamics as the learnable property of a video model itself, without the explicit solver at inference time.

\myparagraph{Physics-Aware Video Synthesis.}
Recent video diffusion models exhibit strong motion priors but are not necessarily physically faithful, and even frontier models still violate basic physical principles in scenes with complex dynamics~\cite{motamed2025generative,bansal2025videophy}. A line of work combines traditional simulation with video generation. PhysGen~\cite{liu2024physgen} couples video generation with rigid-body physics through SDEdit~\cite{meng2021sdedit}, PhysMotion~\cite{tan2024physmotion} extends this into the elastic domain, and PhysDreamer~\cite{zhang2024physdreamer} distills physical parameters using a video model as prior. These methods focus on solid materials and cannot naturally extend to volumetric, turbulent fluids. A second family~\cite{zhang2025think} instead steers diffusion sampling with LLM-derived physical context. Another line of work fine-tunes video models on labeled synthetic data to emit auxiliary channels alongside RGB, though these channels are appearance properties, \eg intrinsics~\cite{zeng2024rgbx,liang2025diffusionrenderer} or weather~\cite{lin2025controllable}, rather than the dynamical state we predict.

To accommodate fluid effect synthesis, a recent line of work~\cite{li2025wonderplay,zhan2026perpetualwonder,liu2026realwonder} built the system on top of the multi-physics simulator~\cite{Genesis}, conditioning the video generation model through bi-modal control. Although these hybrid systems improve physical plausibility for fluids and also other materials, they still require an authored 3D scene and a configured simulator, limiting their applicability to generation from in-the-wild content. We share their goal but bake the physics directly into the video model through our model design and the training pipeline, with no manual effort on explicit simulators.

\myparagraph{Preference Optimization Post-Training.}
Preference optimization, as a post-training stage, has become a standard recipe for aligning generative models with human-defined high-dimensional quality. DPO~\cite{rafailov2023direct} provides a simple offline formulation for the optimization target, and Diffusion-DPO~\cite{wallace2024diffusion} further adapts it to training with diffusion models. VideoDPO~\cite{liu2025videodpo} extends preference alignment to the video diffusion domain with a multi-aspect scoring system, and DenseDPO~\cite{wu2026densedpo} introduces fine-grained temporal preference labels to better target motion quality. EpipolarDPO~\cite{kupyn2025epipolar} employs epipolar constraints to enhance geometry consistency of video generation. Inspired by these, we further treat physical quality, especially complex fluid effects, as another signal that can be used for preference optimization. In detail, we leverage our model's native velocity field prediction as part of the reward signal, turning an auxiliary physical output into a physics-grounded preference indicator at the alignment stage.

\section{Method}
\label{sec:method}

\begin{figure*}[t!]
  \centering
  \includegraphics[width=\linewidth]{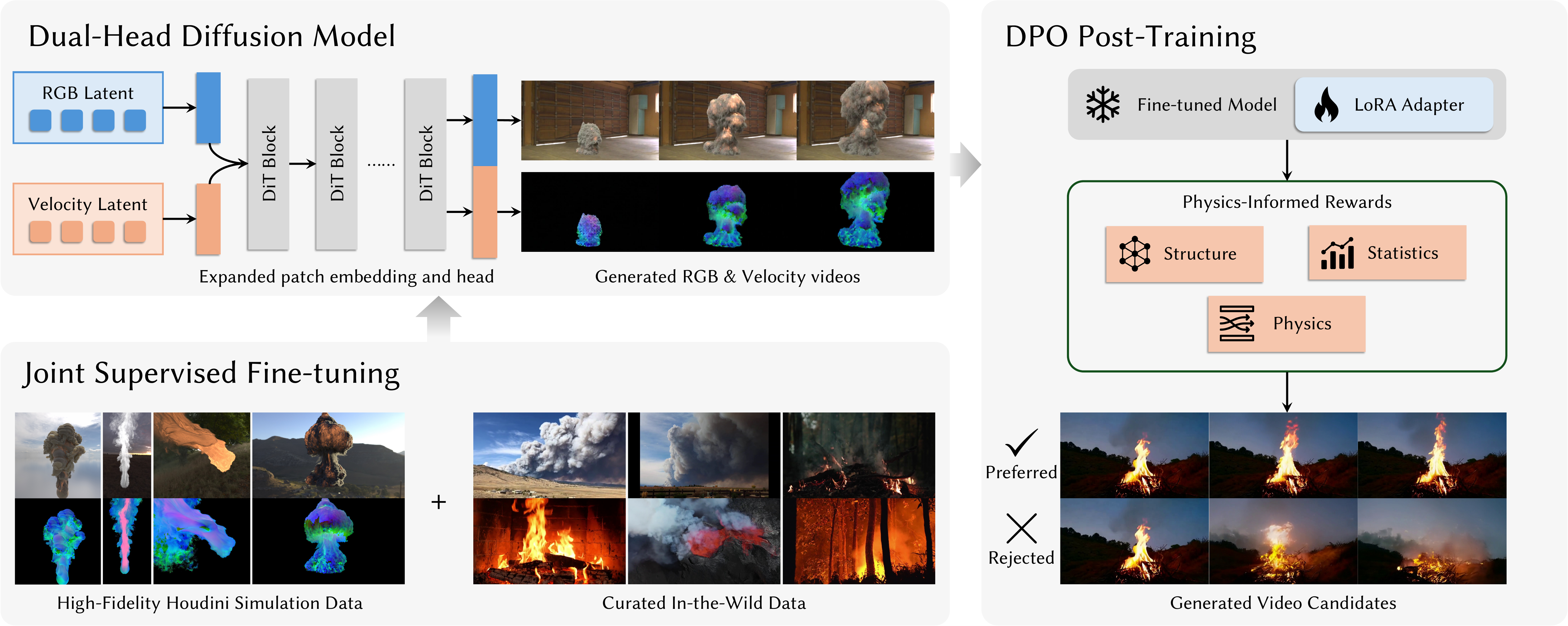}
  \vspace{-10pt}
  \caption{\textbf{Overview of FluidGen}.
  Our pipeline augments a pretrained video diffusion model with a velocity prediction stream and train it in two stages: joint supervised fine-tuning on simulated and in-the-wild fluid videos, followed by velocity-aware DPO using physics-informed rewards.
  }
  \vspace{-10pt}
\label{fig:method}
\end{figure*}

\subsection{Overview}
\label{sec:method:overview}

Our model is trained to jointly generate a fluid video and a pixel-aligned 3D velocity field video using a single video diffusion model, and subsequently trained with the reward metric computed directly from its own velocity field output. Therefore, velocity and RGB predictions serve two key roles. First, as an auxiliary motion signal that  regularizes the appearance stream toward more coherent and consistent dynamics through this joint prediction~\cite{chefer2025videojam}.
More importantly, it transforms every generated sample into the physical dynamics that can be evaluated directly, thereby enabling a velocity-based preference optimization stage without requiring any external motion estimator or simulator.

Concretely, our architecture is a dual-head extension of a pretrained video diffusion transformer~(DiT~\cite{peebles2023scalable})~(\S\ref{sec:method:arch}) that produces both an RGB stream and a velocity stream from a shared latent. The model is trained in two stages: 1)~Joint supervised fine-tuning~(\S\ref{sec:method:sft}): a flow-matching~\cite{lipman2022flow} objective over the joint latent trained on a mixed corpus of Houdini simulations with paired ground-truth velocity, and in-the-wild fluid videos. 2)~Velocity-aware DPO post-training~(\S\ref{sec:method:dpo}): a DPO objective driven by a velocity-informed reward that evaluates each generated trajectory using its predicted velocity field.
Both stages share the same model and output velocity representation, which we introduce first in \S\ref{sec:method:vel}.

\subsection{Velocity Field Representation}
\label{sec:method:vel}

At every pixel of each generated frame, we represent the 3D motion of the underlying fluid particles as a velocity vector $\mathbf{v}=(v_x,v_y,v_z)\in\mathbb{R}^{3}$, expressed in the camera coordinate system. To remain compatible with the standard video model latent space and preserve the pretrained VAE encoder without modification, we do not regress $\mathbf{v}$ directly. Instead, we encode $\mathbf{v}$ into a three-channel color $\mathbf{c}\in[0,1]^{3}$ via a spherical decomposition, as illustrated in Fig.~\ref{fig:dataset_visualization}, and have the velocity head predict $\mathbf{c}$ as an auxiliary RGB-like stream. Let $\rho=\|\mathbf{v}\|$, $\theta=\arccos(v_z/\rho)\in[0,\pi]$, and $\phi=\mathrm{atan2}(v_y,v_x)\in[-\pi,\pi]$. The encoding is defined as
\begin{equation}
\mathbf{c}
=
\Big(
\tfrac{\rho}{S},\;
\tfrac{\theta}{\pi},\;
\tfrac{\phi+\pi}{2\pi}
\Big)
\in[0,1]^{3},
\quad
S
=
\sigma\sqrt{H^{2}+W^{2}},
\label{eq:vel-encode}
\end{equation}
where $(H,W)$ is the image resolution and the global scale $\sigma=10^{-2}$ is chosen such that typical advection magnitudes occupy the middle of the $[0,1]$ range. At inference, the predicted $\mathbf{c}$ is decoded back to $\mathbf{v}$ by inverting Eq.~\ref{eq:vel-encode}.

\subsection{Dual-Head Joint Diffusion Architecture}
\label{sec:method:arch}

Our model builds upon a pretrained 14B image-to-video Diffusion Transformer~\cite{wan2025wan}, whose VAE encodes a video into a $16$-channel spatiotemporal latent $\mathbf{z}^{\mathrm{rgb}} \in\mathbb{R}^{16\times T\times h\times w}$.
We extend the backbone with a parallel \emph{velocity head} operating on a velocity latent of identical shape, and concatenate the two along the channel dimension to form a joint latent, following VideoJam~\cite{chefer2025videojam}:
\begin{equation}
\tilde{\mathbf z}
=
[\mathbf z^{\mathrm{rgb}};\,\mathbf z^{\mathrm{vel}}]
\in\mathbb{R}^{32\times T\times h\times w}.
\end{equation}
The velocity latent is encoded and decoded using the same VAE as the RGB stream: we simply treat the encoded velocity image~(Eq.~\ref{eq:vel-encode}) as an additional RGB clip. This preserves the pretrained perceptual capability of the backbone while avoiding the need to train a separate encoder-decoder latent space.
Two transformer components are widened to support the joint latent~(Fig.~\ref{fig:method}):

\myparagraph{Patch Embedding Input.}
The initial patchify projection is expanded so that its input depth grows from $16+K$ to $16+16+K$, where $K$ denotes the number of conditioning channels~(conditional image and binary mask latent). The first $16+K$ channels reuse the pretrained weights directly, while the additional $16$ corresponding to the velocity latent are zero-initialized. So the model can start the training stage without losing the base prediction ability.

\myparagraph{Output Head.}
The final projection layer mapping tokens back to patch space is widened from an output dimension of $16$ to $32$. The RGB channels reuse the pretrained weights, while the velocity channels are initialized to zero.
This zero initialization scheme preserves the pretrained model behavior at initialization: the RGB branch is initially identical to the original pretrained model, while the velocity branch remains inactive at the beginning of the joint supervised fine-tuning. We empirically explored architectures with separate denoisers and explicit cross-attention bridges, but found that a single shared denoiser sampled along one trajectory already maintains strong causal consistency between RGB and velocity while being substantially simpler for training and inference.

\subsection{Joint Supervised Fine-Tuning}
\label{sec:method:sft}

\begin{figure}[t!]
  \centering
  \includegraphics[width=\linewidth]{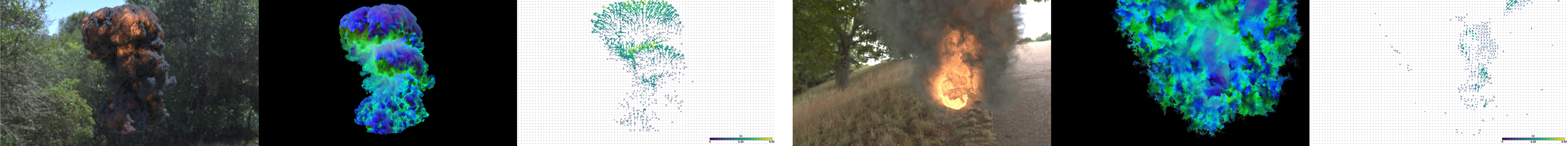}
  \vspace{-10pt}
  \caption{
  \textbf{Data visualization.}
 Here we show two examples of simulated scenes: the rendered RGB frames, velocity field frames and visualized velocity directions in our Houdini-constructed dataset.
    }
\vspace{-10pt}
\label{fig:dataset_visualization}
\end{figure}

\myparagraph{Training Data.}
The dual-head model is fine-tuned on two complementary datasets.
1)~High-fidelity Houdini simulations of smoke, fire, and explosion scenes rendered together with dense ground-truth Eulerian velocity fields, which provide clean $(\text{RGB},\mathbf v)$ supervision as illustrated in Fig.~\ref{fig:dataset_visualization} and anchor the velocity head to physically plausible fluid behavior. Please refer to \S\ref{sec:experiments} and Appendix~\ref{sec:supp_dataset} for more details on the dataset construction.
2)~A curated collection of in-the-wild fluid videos supplies appearance diversity, including lighting, materials, scales, and camera motion, which is difficult to produce in simulation and is critical for generalization to real-world inputs.

\myparagraph{Joint Flow-Matching Objective.}
Both heads are trained jointly with a single flow-matching objective~\cite{lipman2022flow,wan2025wan} over the combined $32$-channel latent. Let $\mathbf z_{1}=[\mathbf z_{1}^{\mathrm{rgb}}; \mathbf z_{1}^{\mathrm{vel}}]$ denote the encoded clean target, let $\mathbf z_{0}\sim\mathcal{N}(0,\mathbf I)$ be the noise sampled from a standard Gaussian distribution, and let $\mathbf z_{t}=(1-t)\mathbf z_{0}+t\mathbf z_{1}$ be the linear interpolation. The denoiser $\mathbf v_{\Theta}(\mathbf z_{t},t,\mathbf c,\mathbf y)$ is trained to predict the conditional flow $\mathbf z_{1}-\mathbf z_{0}$ via
\begin{equation}
\mathcal{L}_{\mathrm{SFT}}
=
\mathbb{E}_{t,\,\mathbf z_{0},\,\mathbf z_{1}}
\Big[
m\,\big\|
\mathbf v_{\Theta}(\mathbf z_{t},t,\mathbf c,\mathbf y)
-(\mathbf z_{1}-\mathbf z_{0})
\big\|^{2}
\Big],
\label{eq:sft}
\end{equation}
where $\mathbf c$ denotes the image-conditioning input and $\mathbf y$ the text prompt. The per-channel mask $m$ equals $1$ for all channels on samples with velocity supervision (the Houdini subset), and equals $0$ only on RGB channels for samples without velocity targets (the in-the-wild subset). Consequently, the velocity head is trained exclusively on simulation data, while the RGB head benefits from the full appearance distribution. In practice, the two datasets are interleaved at a fixed mixing ratio during training.

\subsection{Velocity-Aware DPO Post-Training}
\label{sec:method:dpo}

First stage fine-tuning yields a model capable of jointly generating an RGB stream and a plausible velocity field stream in a single forward pass. However, the physical quality of generated samples still exhibits suboptimal performance. Many samples will remain visibly degenerate, as it is hard for the model to simply learn the physical process~\cite{bansal2025videophy} -- especially for highly unconstrained and complex fluid dynamics.
Consequently, we employ a post-training stage to reduce this variability by aligning the generator with the physics-informed reward, through the preference optimization.

\myparagraph{Physics-Informed Reward.}
\label{sec:method:reward}
We assign each generated sample a scalar reward
$R(\tilde{\mathbf z})\in[0,1]$, defined as a convex combination of thirteen component rewards. Each component acts as a soft indicator of physical plausibility. Eleven components are computed directly from the decoded velocity field and fall into three categories:

\noindent\textit{Structural} terms: including direction diversity, temporal coherence, and multi-scale gradient detail, which encourage non-trivial and temporally stable flow structure;

\noindent\textit{Statistical} terms: including activity naturalness, temporal dynamics balance, life-cycle envelope, and radial outflow, which encourage agreement with the empirical distribution of Houdini simulations;

\noindent\textit{Physical} terms: including advection self-consistency, Q-criterion diversity, direction-field enstrophy~\cite{hunt1988eddies}, and Kolmogorov inertial range compliance~\cite{kraichnan1974kolmogorov}, which encourage classical properties of turbulent flow.

\begin{figure}
  \centering
  \includegraphics[width=\linewidth]{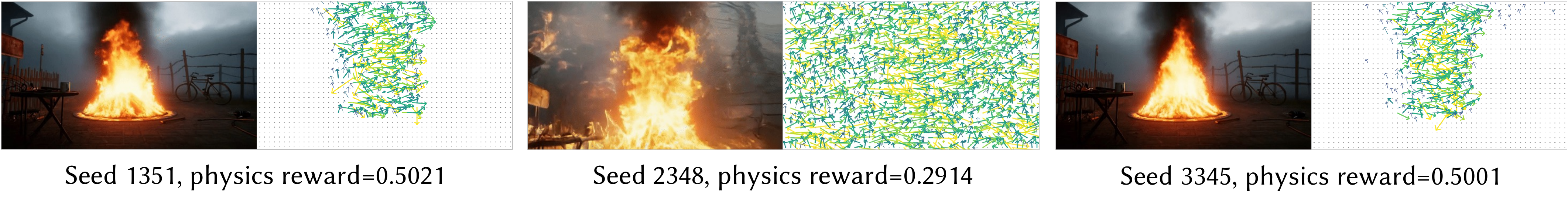}
  \vspace{-20pt}
  \caption{\textbf{Physics-informed reward} distinguishes plausible from implausible fluid dynamics. Four samples~(left: RGB frame; right: visualized velocity field) from the same input annotated with their reward values. High reward samples exhibit organized, turbulent flow structure with rich directional diversity and coherent vortical motion, while low reward samples show collapsed velocity fields that do not correspond to realistic fluid behavior.}
\label{fig:physics reward}
\vspace{-10pt}
\end{figure}

As illustrated in Fig.~\ref{fig:physics reward}, this composite reward reliably separates faithful fluid dynamics from unrealistic ones, and serves as the criterion for preference pairs in the post-training stage.

\myparagraph{Preference Pairs and Flow-DPO Objective.}
\label{sec:method:flowdpo}
For each conditioning pair $(\mathbf c_i,\mathbf y_i)$ drawn from the first stage training corpus, we sample $N{=}4$ candidates using independent random seeds. Since each sample is itself a $32$-channel video latent containing both RGB and velocity streams generated by the same denoiser, the aforementioned reward can be evaluated directly on the model output without any additional motion estimator. Within each candidate set, we construct a preference pair $(\tilde{\mathbf z}^{+},\tilde{\mathbf z}^{-})$ using the highest and lowest reward samples, respectively, and discard pairs whose reward gap is below a threshold $\delta$.

Given a pair at diffusion time $t$, let $\mathbf u^{\star}_{t}$ denote the flow-matching target from Eq.~\ref{eq:sft}, and let $\mathbf v_{\Theta}$ and $\mathbf v_{\Theta_{\mathrm{ref}}}$ denote the trainable and reference denoisers, respectively. Following the MSE-based Diffusion-DPO formulation~\cite{wallace2024diffusion} adapted to flow matching~\cite{rafailov2023direct,liu2025videodpo}, we define the per-sample preference residual as:
\begin{equation}
\Delta(\tilde{\mathbf z})
=
\big\|
\mathbf v_{\Theta}(\tilde{\mathbf z},t)-\mathbf u^{\star}_{t}
\big\|^{2}
-
\big\|
\mathbf v_{\Theta_{\mathrm{ref}}}(\tilde{\mathbf z},t)-\mathbf u^{\star}_{t}
\big\|^{2},
\label{eq:dpo-delta}
\end{equation}
and optimize
\begin{equation}
\mathcal{L}_{\mathrm{DPO}}
=
-
\mathbb{E}_{(\tilde{\mathbf z}^{+},\tilde{\mathbf z}^{-}),\,t}
\;
\log\sigma\ \Big(
-\tfrac{1}{2}\beta
\big(
\Delta(\tilde{\mathbf z}^{+})
-
\Delta(\tilde{\mathbf z}^{-})
\big)
\Big),
\label{eq:dpo-loss}
\end{equation}
where $\sigma(\cdot)$ denotes the logistic sigmoid function and $\beta$ is a temperature parameter. The objective decreases when the trainable model improves its prediction error on preferred samples more than on dispreferred samples, relative to the reference model. During this stage, we optimize $\mathbf v_{\Theta}$ using low-rank adaptation~(LoRA~\cite{hu2022lora}) applied to attention and FFN linear layers while freezing the backbone parameters. The reference model $\mathbf v_{\Theta_{\mathrm{ref}}}$ is obtained implicitly by disabling the LoRA adapters during inference.

\section{Experiments and Results}
\label{sec:experiments}

We evaluate our method's ability to generate visually compelling and physically plausible fluid phenomena. Our experiments are designed to answer three key questions: 1) Does our dual-head architecture and training pipeline improve the physical fidelity of generated fluids compared to state-of-the-art video models? 2) How do the joint velocity prediction and the subsequent DPO post-training individually contribute to the final performance? Please refer to our project page for more video results and comparisons.

\begin{figure*}[t!]
  \centering
  \includegraphics[width=\linewidth]{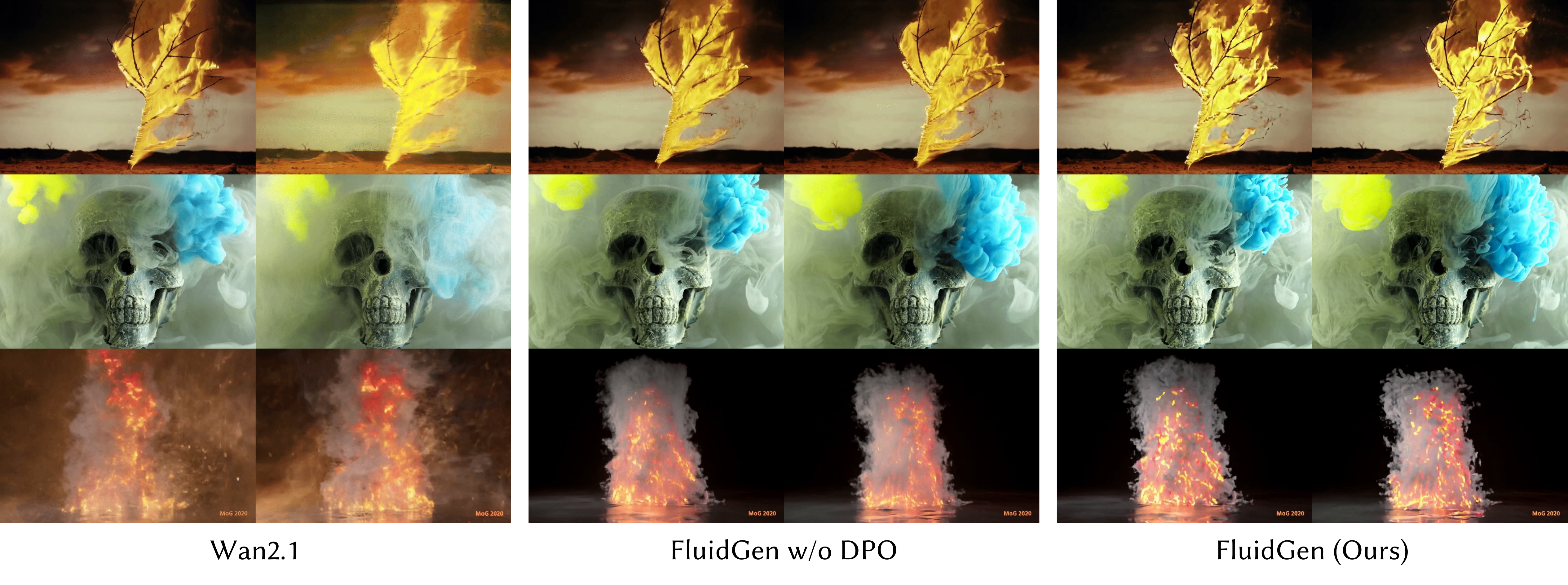}
  \vspace{-20pt}
  \caption{\textbf{Qualitative comparison on in-the-wild inputs}. Wan2.1-I2V-14B produces visibly unnatural fluid motion with appearance drift,
  while the model after the Joint Supervised Fine-tuning (\moniker \wo DPO) generates physically coherent dynamics.
  Our full solution with DPO (\moniker) further sharpens the result, yielding richer fine-scale details and more vivid fire and smoke motion. Please refer to our project page for better visualization.}
\label{fig:FT DPO}
\end{figure*}

\begin{figure}
  \centering
  \includegraphics[width=\linewidth]{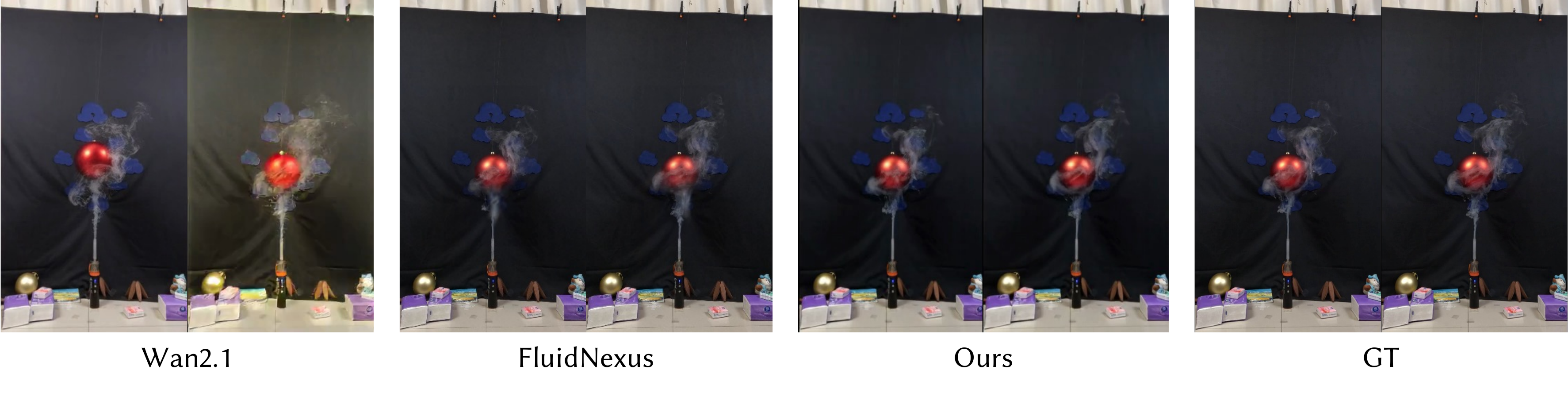}
  \vspace{-20pt}
  \caption{
 \textbf{Qualitative comparison on the FluidNexus~\cite{gao2025fluidnexus} dataset}.
 Even with costly multi-view reconstruction, FluidNexus
 still omits some of the fine details,
 while the pretrained Wan2.1 fails to generate realistic fluid dynamics and
 consistent scene appearance.
    }
\vspace{-10pt}
\label{fig:fluidnexus}
\end{figure}

\subsection{Experimental Setup}
\myparagraph{Datasets.}
As aforementioned in \S\ref{sec:method:sft}, our training dataset is consist of two sources. In detail, we constructed $13125$ videos from Houdini simulation, 15570 real-world videos from FluidNexus dataset~\cite{gao2025fluidnexus}, and curated $6521$ videos from the real-world footage. In total resulting in a training set of more than $35$K videos. Each video is resized to the desired $480\times 832$ resolution, for portrait videos we equally pad black pixels on left and right to match the size ratio. Each video has $81$ frames for training.

\myparagraph{Implementation Details.}
Our model is initialized from the pretrained Wan-2.1-14B image-to-video diffusion model~\cite{wan2025wan}. During the joint Supervised Fine-Tuning~(FT) stage, we train the widened transformer blocks for 7000 iterations with a batch size of 1 and a learning rate of 1e-5. For the Velocity-Aware DPO stage, we freeze the backbone and train LoRA adapters~(rank $r=64$) for 500 iterations using a preference margin threshold $\delta=0.05$ and temperature $\beta=500$. Training was conducted on 8 NVIDIA H100 GPUs. The codebase is built on DiffSynth~\cite{diffsynth_studio} and Epipolar-DPO~\cite{kupyn2025epipolar}.

\myparagraph{Baselines.}
We compare our approach against strong foundational video generation models and recent physics-aware methods:
1) \textbf{Wan-14B}~\cite{wan2025wan}: The state-of-the-art base model our architecture is built upon, serving as the direct baseline for our physical augmentations. This model also stands for the current performance of the latest open-source video model can achieve on fluid dynamics.
2) \textbf{FluidNexus}~\cite{gao2025fluidnexus}: On benchmarks with ground-truth data, we also compare with an \textit{upgraded multi-view version} of the original FluidNexus model. We directly use applicable multi-view videos to reconstruct the underlying fluid representations. And show that even with the additional multi-view information, the pure physics-based approach still can't faithfully generate complex fluid dynamics at the desired level.

\begin{table*}[t!]
    \centering
    \setlength{\tabcolsep}{4pt}
    \renewcommand{\arraystretch}{1.05}
    \begin{minipage}[t]{0.48\textwidth}
        \centering
        \caption{Quantitative results on the FluidNexus~\cite{gao2025fluidnexus} dataset.}
        \label{tab:fluidnexus}
        \small
        \begin{adjustbox}{max width=\linewidth}
        \begin{tabular}{lcccc}
            \toprule
            Method & PSNR$\uparrow$ & SSIM$\uparrow$ & LPIPS$\downarrow$ & FVD$\downarrow$ \\
            \midrule
            FluidNexus     & 31.04 & 0.9040 & 0.1014 & 542.6 \\
            Wan2.1-I2V-14B & 22.50 & 0.8018 & 0.1821 & 569.7 \\
            \midrule
            \moniker\ \wo\ DPO & \textbf{31.52} & \textbf{0.9203} & 0.0547 & \textbf{192.1} \\
            \moniker\ (Ours)   & 31.08 & 0.9157 & \textbf{0.0542} & 199.2 \\
            \bottomrule
        \end{tabular}
        \end{adjustbox}
    \end{minipage}
    \hfill
    \begin{minipage}[t]{0.48\textwidth}
        \centering
        \caption{VBench~\cite{huang2024vbench} scores for image-to-video generation on in-the-wild data.}
        \label{tab:vbench}
        \small
        \begin{adjustbox}{max width=\linewidth}
        \begin{tabular}{lccc}
            \toprule
            Method & Quality$\uparrow$ & Aesthetics$\uparrow$ & Consistency$\uparrow$ \\
            \midrule
            Wan2.1-I2V-14B & 0.4332 & 0.5144 & 0.8929 \\
            \midrule
            \moniker\ \wo\ DPO & 0.4282 & 0.5255 & \textbf{0.9383} \\
            \moniker\ (Ours)   & \textbf{0.4627} & \textbf{0.5283} & 0.9350 \\
            \bottomrule
        \end{tabular}
        \end{adjustbox}
    \end{minipage}
    \vspace{-10pt}
\end{table*}

\myparagraph{Metrics.} We evaluate along two axes using two benchmarks. The \emph{FluidNexus} benchmark~\cite{gao2025fluidnexus} provides multiview captures of real fluids with ground-truth reference frames; on $15$ randomly selected sequences, we report PSNR and SSIM for per-frame reconstruction fidelity, LPIPS~\cite{zhang2018unreasonable} for perceptual similarity, and FVD~\cite{unterthiner2019fvd} for distributional distance over spatiotemporal features. To evaluate general video quality on in-the-wild image inputs, we additionally report VBench~\cite{huang2024vbench} scores on $20$ selected images along three dimensions relevant to fluid synthesis: Imaging Quality (per-frame visual fidelity), Aesthetic Quality (perceptual appeal), and Subject Consistency (temporal stability of foreground content, critical for coherent advection without flickering).

\subsection{Quantitative Evaluation}

Tab.~\ref{tab:fluidnexus} reports results on the FluidNexus benchmark. We note that the FluidNexus baseline used here is an upgraded multi-view variant of the original method, which consumes multiple synchronized views as input and requires several hours of per-scene optimization to reconstruct the underlying fluid representation. Despite this substantially heavier setup, it still fails to recover some fine level turbulent details such as small scale eddies and high-frequency boundary roll-ups, which is reflected in its lower SSIM and notably higher LPIPS and FVD compared to our method. Wan2.1-I2V-14B, on the other hand, struggles to produce faithful fluid dynamics from a single image and frequently introduces appearance artifacts such as color shifting and texture drift over time. In contrast, our model, conditioned on only a single image and run in a single forward pass, achieves the best or on par metrics with the multi-view FluidNexus baseline, indicating that our generated complex fluid dynamics closely match the ground-truth captures.

To verify that our training stages do not compromise general generative capability on in-the-wild content, we evaluate our method using VBench metrics in Tab.~\ref{tab:vbench}. Our final model matches or surpasses the Wan2.1-I2V-14B base model on all three reported dimensions, with clear gains in Imaging Quality and Subject Consistency. The improvement in Subject Consistency is particularly meaningful for fluid scenes: it indicates that the auxiliary velocity stream regularizes the RGB stream toward temporally coherent advection, mitigating the flickering and shifting artifacts observed in the base model. Together, the two benchmarks show that our method substantially improves physical fidelity.

\subsection{Qualitative Results}

\begin{figure*}[!t]
  \centering
  \includegraphics[width=0.95\linewidth]{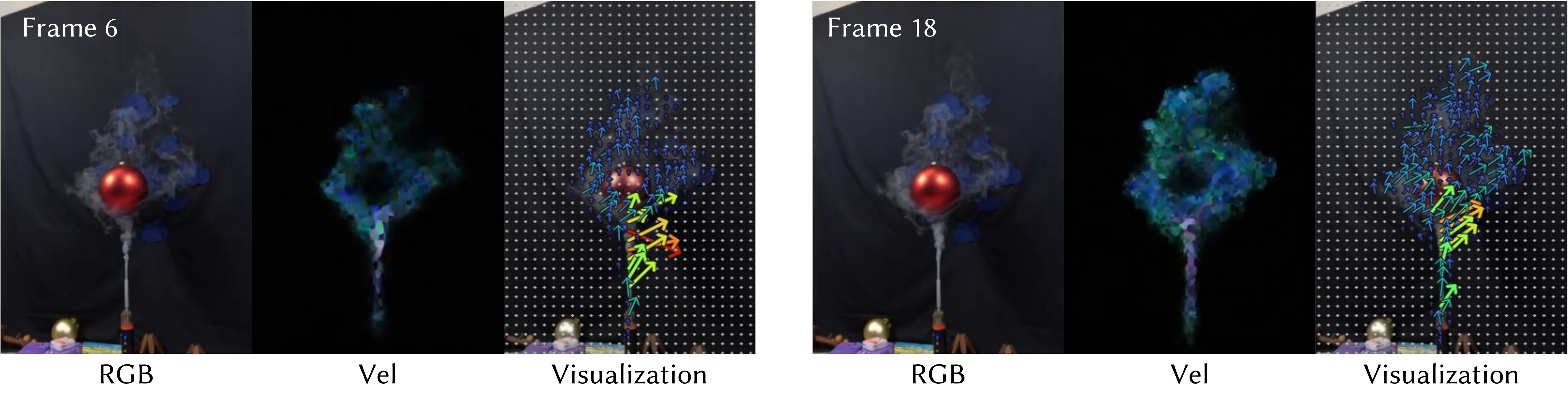}
  \vspace{-10pt}
  \caption{\textbf{Generated velocity fields} on a real-world capture. Here we show the input RGB (\textit{left}), the generated velocity encoded as color (\textit{middle}, Eq.~\ref{eq:vel-encode}), and the decoded velocity overlaid as arrows (\textit{right}). The recovered field reveals coherent advection consistent with the observed smoke motion.}
  \vspace{-10pt}
\label{fig:vel}
\end{figure*}

Fig.~\ref{fig:fluidnexus} shows visual comparisons against ground-truth captures from the FluidNexus benchmark. The upgraded FluidNexus baseline recovers the coarse shape and trajectory of the smoke plume, but its reconstructions after several hours of optimization still appear overly smooth and lose some fine-scale wisps and turbulent filaments visible in the ground truth, indicating the disadvantage of relying only on the manually defined physical approximation laws.
Wan2.1-I2V-14B produces motion that diverges from the reference dynamics and exhibits noticeable color drift and appearance inconsistency over time. In contrast, our method reproduces both the global rising trajectory and fine turbulent structures, including thin tendrils at the plume boundary, while requiring only a single forward pass from a single input image.
We further compare against Wan2.1-I2V-14B on in-the-wild inputs in Fig.~\ref{fig:FT DPO}, where our full model produces more vivid and physically realistic fluid dynamics. Finally, our model supports several downstream applications, including fluid-related visual effects~(Fig.~\ref{fig:teaser}) via keyframe control~(images are generated by NanoBanana 2~\cite{nanobanana2_2026}), velocity field prediction~(Fig.~\ref{fig:vel}) from input videos using RePaint-style~\cite{lugmayr2022repaint} conditional sampling, and fluid simulation driven by the generated 3D velocity fields~(Fig.~\ref{fig:fluid_sim}). More results are provided in Appendix~\ref{sec:supp_results} and on our project page.

\begin{figure*}[!t]
  \centering
  \includegraphics[width=0.95\linewidth]{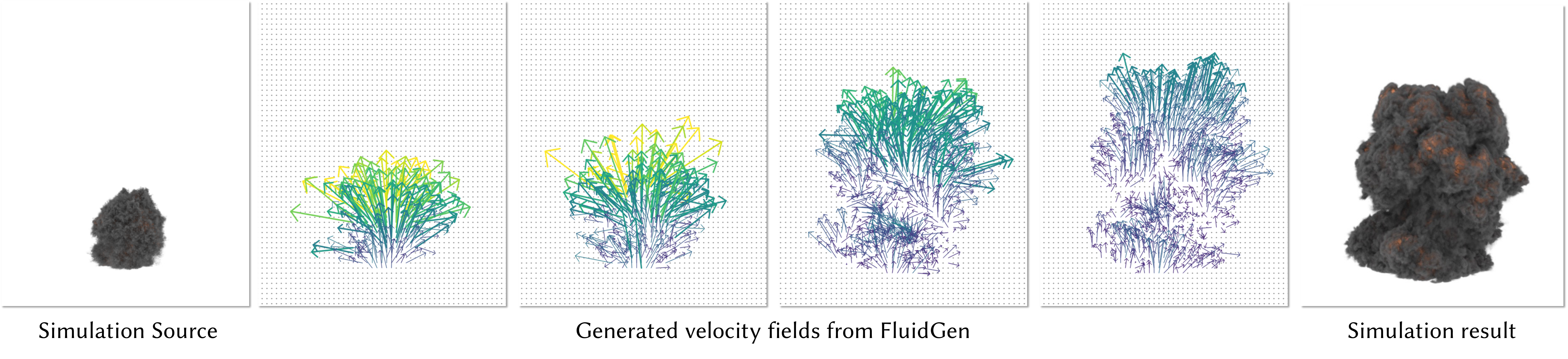}
  \vspace{-10pt}
  \caption{\textbf{Fluid simulation from generated velocity fields}. Our generated fluid velocity fields are physically accurate and can be directly employed in Houdini to simulate a fluid source.}
  \vspace{-10pt}
\label{fig:fluid_sim}
\end{figure*}

\subsection{Ablation Studies}
\label{sec:experiments:ablation}
We ablate the two stages of our pipeline against the Wan2.1-I2V-14B base model. Results are reported in Tab.~\ref{tab:fluidnexus} and Tab.~\ref{tab:vbench}; qualitative comparisons in Fig.~\ref{fig:FT DPO}.

\myparagraph{Joint Supervised Fine-Tuning.}
The stage I fine-tuning achieves significant gain compared to the basemodel: \moniker \wo DPO improves PSNR from $22.50$ to $31.52$ and reduces FVD from $569.7$ to $192.1$ over the base model, while improving Subject Consistency on VBench ($0.8929 \rightarrow 0.9383$).
We attribute this improvement to our curated training corpus of high-fidelity Houdini simulations and in-the-wild fluid videos, which exposes the model to the turbulent dynamics largely absent from the base model's pretraining distribution.

\myparagraph{Velocity-Aware DPO.}
DPO trades a small amount of reconstruction fidelity for noticeably better perceptual quality. FluidNexus metrics remain on par with the FT-only model, while VBench Imaging Quality improves significantly and Aesthetic Quality slightly. The slight drop in Subject Consistency ($0.9383 \rightarrow 0.9350$) is expected: the preference signal favors higher-energy, more turbulent motion. Qualitatively (Fig.~\ref{fig:FT DPO}), the FT+DPO model (\moniker) produces richer fine-scale details and more vivid fire and smoke motion than \moniker \wo DPO. Please refer to our project page for better visualization.

\section{Conclusion}

We present FluidGen, a framework for synthesizing physically plausible fluid dynamics directly within a video diffusion model, bridging the gap between the rich appearance priors of generative models and the dynamic constraints of classical fluid solvers. By augmenting a pretrained video diffusion transformer with a dual-head architecture, our model jointly predicts RGB pixels and an underlying 3D velocity field, capturing complex fluid dynamics without authored 3D scenes or explicit simulators at inference.

Our two-stage training pipeline first jointly fine-tunes the model with dense physical supervision from high-fidelity Houdini simulations alongside the visual diversity of in-the-wild videos, then applies a velocity-aware preference optimization stage that uses the model's native velocity predictions to compute a physics-informed reward over structural, statistical, and fluid properties, aligning the generator toward turbulent, coherent, and physically sound motion.

While our method focuses on volumetric fluids such as smoke, fire, and explosions, the core idea of embedding interpretable physical representations into generative latents points toward a broader paradigm for physics-aware synthesis. Future work could extend the dual-head approach to other physical systems such as liquids or elastic materials, or expose the predicted velocity fields as a direct control handle for artists. Our results demonstrate that with the right internal representations and alignment strategies, large-scale video models can natively master complex physical dynamics.

\bibliography{main}
\bibliographystyle{iclr2027_conference}

\clearpage
\appendix
\section{Additional Results}
\label{sec:supp_results}

\begin{figure*}[!t]
  \centering
  \includegraphics[width=0.95\linewidth]{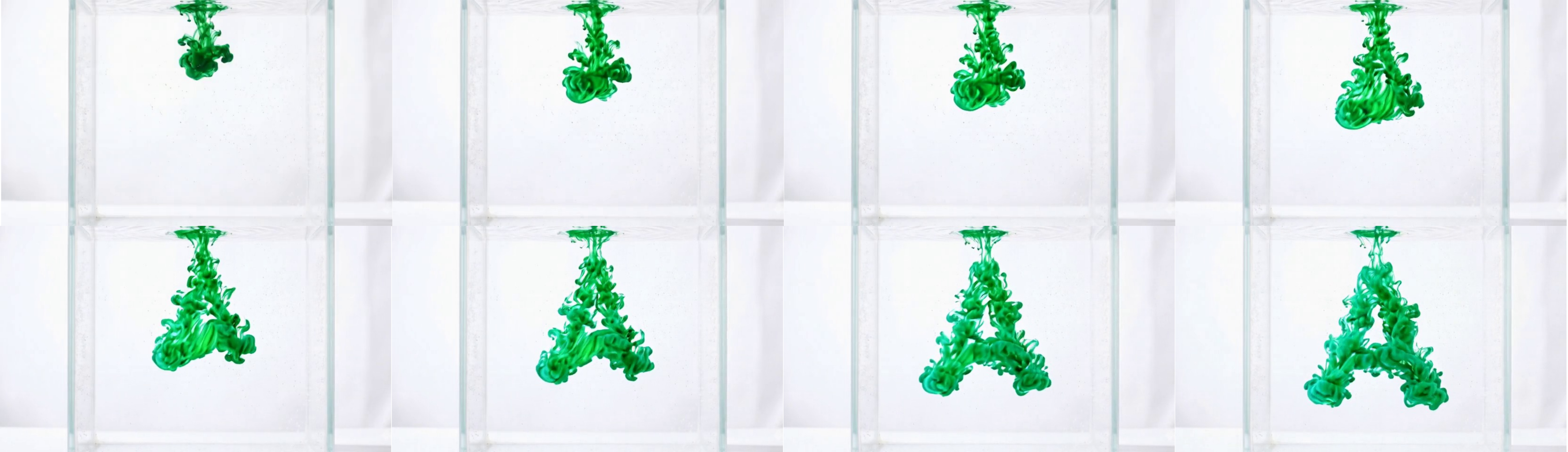}
  \caption{\textbf{Visual effects generation} using FluidGen.
  The two rows are ordered frames from a video generated by FluidGen.
  Given keyframes, FluidGen generates a realistic fluid dynamic transition, here ink in water forming the letter ``A''.
  }
\label{fig:vfx}
\end{figure*}

\begin{figure*}[!t]
  \centering
  \includegraphics[width=\linewidth]{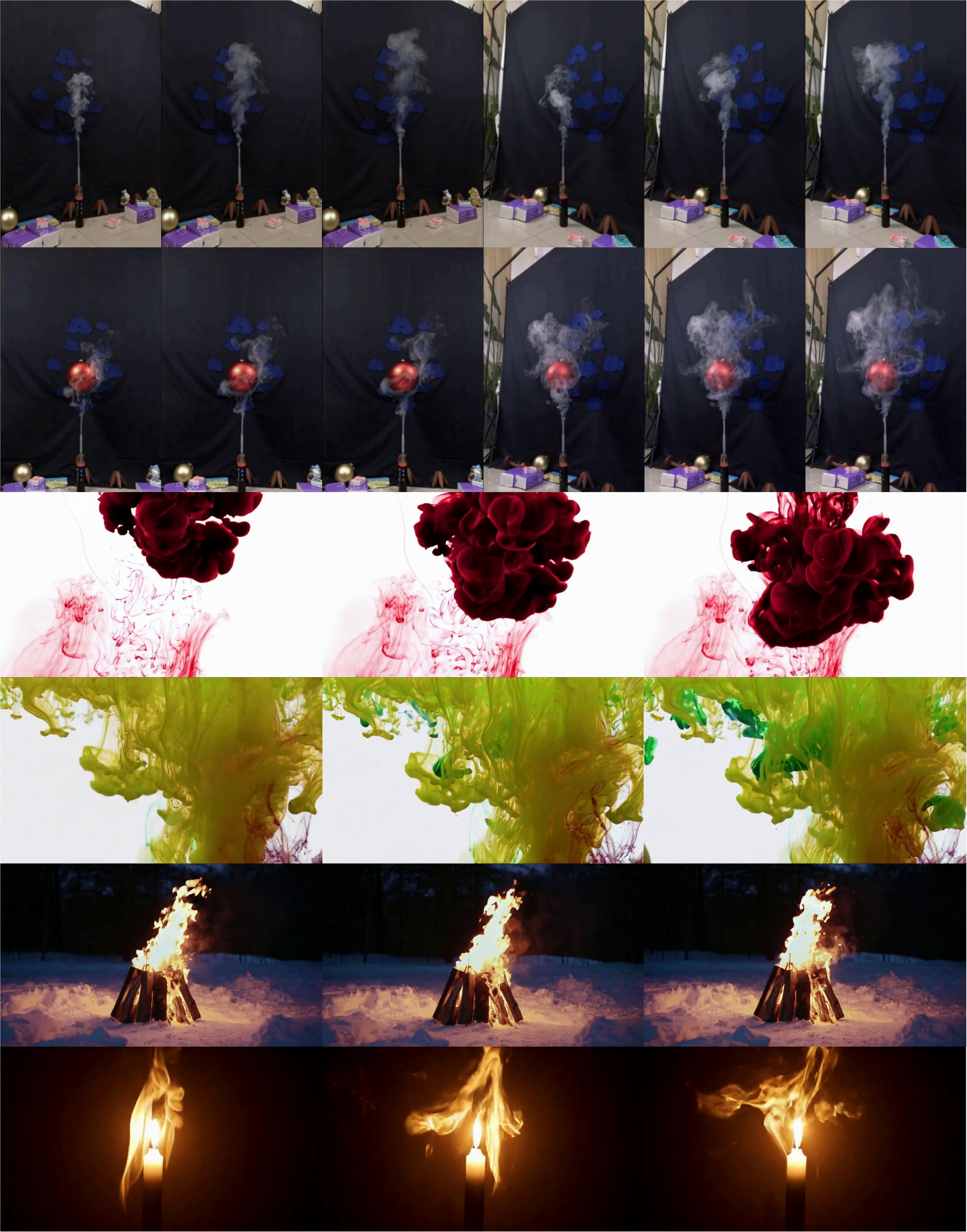}
  \caption{\textbf{FluidGen qualitative results}.
  Each row contains frames from a video generated by FluidGen.
  FluidGen can generate realistic fluid, such as smoke, ink, and fire, in various scenarios.}
\label{fig:add}
\end{figure*}

Fig.~\ref{fig:vfx} shows a keyframe-controlled visual effect generated by \moniker, and Fig.~\ref{fig:add} shows additional qualitative results on smoke, ink, and fire.

\section{More Details}

\subsection{Physics-Informed Reward Components}
\label{sec:supp_rewards}

\mysuppparagraph{Notation.} Let $\mathbf{v}_t(x,y) = (v_x, v_y, v_z)$ denote the per-pixel 3D velocity field at frame $t$ decoded from the velocity stream of a generated sample $\tilde{\mathbf{z}}$ via the inverse transformation of Eq.~\ref{eq:vel-encode}. We write $\rho_t = \|\mathbf{v}_t\|$ for the speed magnitude and $\hat{\mathbf{u}}_t = \mathbf{v}_t / \rho_t$ for the unit-direction field, and set it to zero when $\rho_t$ falls below a small numerical floor value. At each frame we define an active region $\mathcal{A}_t = \{(x,y) : \rho_t(x,y) > \tau_\rho\}$ that filters out near static background pixels, where $\tau_\rho = \alpha_a S$ with $S$ the encoding scale of Eq.~\ref{eq:vel-encode} and $\alpha_a$ a fixed fraction.
We also use the normalized R-channel $R_t \in [0,1]$ of the encoded velocity image directly for temporal-difference statistics. The symbols $\kappa_\bullet$, $\mu_\bullet$, and $\sigma_\bullet$ denote scalar hyper-parameters; their values are calibrated once on a small held-out set of Houdini-simulated and curated in-the-wild clips, and are listed at the end of this section.

\mysuppparagraph{Reward components.}
In total, we compute eleven reward components~(Tab.~\ref{tab:rewards}), which act on the velocity stream and fall into three groups.

Structural terms target the spatial and temporal organization of the direction field:
\begin{itemize}
    \item $R_{\mathrm{div}}$ penalizes direction collapse via the per-component variance of $\hat{\mathbf{u}}$,
    \item $R_{\mathrm{tc}}$ measures the temporal Pearson correlation of the spatial gradients of $\hat{\mathbf{u}}$,
    \item $R_{\mathrm{msd}}$ rewards consistent gradient richness at scales $s\in\{1,2,4\}$ so that the flow shows structured detail at all scales rather than blur or pure noise.
\end{itemize}

Statistical terms match the empirical distribution of Houdini simulations:
\begin{itemize}
    \item $R_{\mathrm{act}}$ on the mean active-pixel fraction $\bar{f}_a = \mathbb{E}_t[|\mathcal{A}_t|/HW]$,
    \item $R_{\mathrm{tdb}}$ on the mean temporal change of $R_t$,
    \item $R_{\mathrm{lc}}$ on the global rise-peak-decay envelope of per-frame energy $e_t = \mathbb{E}_{x,y}[\rho_t]$ via a weighted combination of peak position, monotone rise, monotone decay, second-derivative smoothness, and contrast sub-scores $\{s_k\}$,
    \item $R_{\mathrm{ro}}$ on the speed weighted cosine alignment between the image plane velocity $(v_x,v_z)$ and the radial direction $\mathbf{r}$ from an automatically estimated source, averaged over early frames.
\end{itemize}

Physical terms encode classical criterions of turbulent flow:
\begin{itemize}
    \item $R_{\mathrm{adv}}$ checks one-step advection self-consistency through a bilinear warp $\mathcal{W}$ of $\mathbf{v}_t$ along $(v_x,v_z)$,
    \item $R_Q$ rewards a spread of the Q-criterion of $\hat{\mathbf{u}}$, $Q_{\hat{u}} = \Omega_{12}^2 - (S_{11}^2 + S_{22}^2 + 2 S_{12}^2)$, where $S$ and $\Omega$ are the symmetric and antisymmetric parts of $\nabla\hat{\mathbf{u}}$~\cite{hunt1988eddies},
    \item $R_{\mathrm{ens}}$ rewards the enstrophy of the directional field vorticity $\omega_{\hat{u}} = \partial_x\hat{u}_z - \partial_z\hat{u}_x$ on $\mathcal{A}_t$,
    \item $R_{\mathrm{k}}$ rewards Kolmogorov inertial range scaling of the second order structure function $S_2(r) = \langle \|\mathbf{v}(\mathbf{x}+\mathbf{r}) - \mathbf{v}(\mathbf{x})\|^2 \rangle_{\mathcal{A}}$ along $r\in\{1,2,4,8,16\}$ pixels, fit to a power law $S_2(r)\sim r^\alpha$ and scored by closeness of $\alpha$ to $2/3$~\cite{kraichnan1974kolmogorov}.
\end{itemize}
The total reward used by the DPO objective in Eq.~\ref{eq:dpo-loss} is the convex combination $R(\tilde{\mathbf{z}}) = \sum_i w_i R_i$ with $w_i \ge 0$ and $\sum_i w_i = 1$.

\begin{table}[t]
\centering
\caption{The thirteen component rewards combined into the total
$R(\tilde{\mathbf{z}}) = \sum_i w_i R_i$ used by the velocity-aware DPO
objective. See Sec.~\ref{sec:supp_rewards} for notation and definitions of
$\mathcal{A}_t$, $\hat{\mathbf{u}}$, $e_t$, $S_2(r)$, and the scalar
hyper-parameters.}
\label{tab:rewards}
\small
\setlength{\tabcolsep}{6pt}
\renewcommand{\arraystretch}{1.35}
\begin{tabular}{@{}l l l@{}}
\toprule
Category & Name & Computing formula \\
\midrule
\multirow{3}{*}{Structural}
 & $R_{\mathrm{div}}$ &
   $\displaystyle \mathbb{E}_{t}\!\left[\tanh\!\Big(\tfrac{1}{3}\!\!\sum_{k\in\{x,y,z\}}\!\!\mathrm{Var}\,\hat{u}_k\big|_{\mathcal{A}_t}\big/\kappa_{\mathrm{div}}\Big)\right]$ \\
 & $R_{\mathrm{tc}}$ &
   $\displaystyle \mathbb{E}_{t}\!\left[\big[\mathrm{Pearson}\!\left(\nabla\hat{\mathbf{u}}_t,\,\nabla\hat{\mathbf{u}}_{t+1}\right)\big]_{+}\right]$ \\
 & $R_{\mathrm{msd}}$ &
   $\displaystyle \mathbb{E}_{t}\!\left[0.7\tanh(g_1/\kappa_g) + 0.3\,\exp\!\big(-\tfrac{1}{2}\log^2(g_1/g_{|\mathcal{S}|})\big)\right]$ \\
\midrule
\multirow{4}{*}{Statistical}
 & $R_{\mathrm{act}}$ &
   $\displaystyle \exp\!\Big(-\tfrac{1}{2}\big(\bar{f}_a - \mu_a\big)^2/\sigma_a^2\Big)$ \\
 & $R_{\mathrm{tdb}}$ &
   $\displaystyle \exp\!\Big(-\tfrac{1}{2}\big(\mathbb{E}_t|R_{t+1}-R_t| - \mu_R\big)^2/\sigma_R^2\Big)$ \\
 & $R_{\mathrm{lc}}$ &
   $\displaystyle \sum_{k}\lambda_k\,s_k(e_t),\quad e_t = \mathbb{E}_{x,y}[\rho_t]$ \\
 & $R_{\mathrm{ro}}$ &
   $\displaystyle \mathbb{E}_{t<T_e}\!\left[\frac{\sum_{\mathcal{A}_t}\!\rho_{xz}\,[\cos\!\angle((v_x,v_z),\mathbf{r})]_{+}}{\sum_{\mathcal{A}_t}\!\rho_{xz}}\right]$ \\
\midrule
\multirow{4}{*}{Physical}
 & $R_{\mathrm{adv}}$ &
   $\displaystyle \mathbb{E}_{t}\!\left[\big[\mathrm{Pearson}\!\left(\mathbf{v}_{t+1},\,\mathcal{W}(\mathbf{v}_t)\right)\big|_{\mathcal{A}_t}\big]_{+}\right]$ \\
 & $R_{Q}$ &
   $\displaystyle \mathbb{E}_{t}\!\left[\tanh\!\big(\mathrm{std}(Q_{\hat{u}})\big|_{\mathcal{A}_t}/\kappa_Q\big)\right]$ \\
 & $R_{\mathrm{ens}}$ &
   $\displaystyle \mathbb{E}_{t}\!\left[\tanh\!\big(\langle\omega_{\hat{u}}^{2}\rangle\big|_{\mathcal{A}_t}/\kappa_\varepsilon\big)\right]$ \\
 & $R_{\mathrm{k}}$ &
   $\displaystyle \exp\!\big(-|\alpha - 2/3|/\sigma_\alpha\big),\;\; S_2(r) \sim r^{\alpha}$ \\
\bottomrule
\end{tabular}
\end{table}

\begin{table*}[ht]
\centering
\small
\caption{Per-asset randomization ranges for Houdini simulation data construction. $\mathcal{U}(a,b)$ stands for uniform distribution and
$\mathcal{N}(\mu,\sigma)$ stands for Gaussian distribution.}
\label{tab:dataset_params}
\begin{tabular}{lccc}
\toprule
                        & \textbf{Fire}                     & \textbf{Smoke}                            & \textbf{Explosion}                \\
\midrule
Solver                  & Houdini Pyro      & Houdini Pyro              & Axiom (GPU)                       \\
Buoyancy                & $\mathcal{U}(0.15, 0.3)$          & $\mathcal{U}(0.5, 1.0)$                   & $\mathcal{U}(0.1, 1.0)$           \\
Turbulence              & $\mathcal{U}(0.8, 2.0)$           & $10^{\mathcal{U}(-0.5,-0.05)}$            & $\mathcal{U}(0, 10)$              \\
Disturbance             & $\mathcal{U}(3, 15)$              & $\mathcal{U}(0.1, 0.2)$                   & $\mathcal{U}(0, 1)$               \\
Dissipation             & $0.01$                            & $\mathcal{U}(0.005, 0.05)$                & $\mathcal{U}(0.005, 0.05)$        \\
Divergence scale        & $\sim\!\mathcal{U}(0.10, 0.15)$   & $\sim\!\mathcal{U}(0.10, 0.15)$           & ---                               \\
Confinement             & ---                               & ---                                       & $\mathcal{U}(0, 3)$               \\
Density scale           & $1$                               & $\mathcal{U}(5, 10)$                      & ---                               \\
Temperature scale       & $1$                               & $3^{\mathcal{U}(-1,1)}$                   & ---                               \\
Fire intensity $k_{\text{fire}}$ & $10^{\mathcal{U}(1, 1.6)}$ & ---                                     & ---                               \\
Emitter center          & $\mathcal{N}(\!(0,-0.75,0), \mathbf{I})$ & $\mathcal{N}(\!(0,-5.5,0), \mathrm{diag}(1,0.1,1))$ & origin (preset) \\
Camera $z$              & $\mathcal{U}(6, 10)$              & $\mathcal{U}(6, 10)$                      & $\mathcal{U}(30, 35)$             \\
Wind fields             & $\{0,1\}$, strength $\mathcal{U}(0, 0.8)$ & $\{0,1\}$, strength $\mathcal{U}(0, 0.8)$ & ---                          \\
Env-light intensity     & $\mathcal{U}(0.1, 0.5)$           & $\mathcal{U}(0.1, 0.15)$ + 2 rim lights   & $\mathcal{U}(0.9, 1.5)$           \\
\bottomrule
\end{tabular}
\end{table*}

\begin{algorithm}[t]
\DontPrintSemicolon
\SetKwInOut{Input}{Input}\SetKwInOut{Output}{Output}
\Input{asset type $a \in \{\texttt{fire},\texttt{smoke},\texttt{explosion}\}$, job id $j$, frame count $T$}
\Output{rendered frames, velocity AOV}

$\boldsymbol{\theta} \gets \textsc{SampleParams}(a)$ \tcp*{Table~\ref{tab:dataset_params}}
$\mathbf{c} \gets \textsc{SampleEmitterCenter}(a)$\;
Build emitter SOP chain: \texttt{sphere}$\to$\texttt{attribnoise}$\to$\texttt{pyrosource}$\to$\texttt{volumerasterize}\;
$S \gets \textsc{BuildPyroSolver}(\boldsymbol{\theta})$\;
\If{wind enabled}{
    Append $N \sim \mathcal{U}\{0,1\}$ rasterized wind VDBs as \texttt{vel} sources of $S$\;
}
Patch shader: \texttt{bind(vel)}$\to$\texttt{bind(velocity, export)}\;
$(\mathbf{p}_{\text{cam}}, R_{\text{w2c}}) \gets \textsc{PlaceCamera}(a)$; attach random HDR env light\;
\textbf{Simulate} frames $1\!\ldots\!T$, cache VDBs to disk\;
\textbf{Render} each cached frame with Karma; emit \texttt{(RGB, velocity)} subimages\;
\ForEach{frame}{
    Read $\mathbf{v}_{\text{world}}$ from the velocity AOV\;
    $\mathbf{v} \gets R_{\text{w2c}}\, \mathbf{v}_{\text{world}}$\;
    Encode $\mathbf{v}$ via spherical coordinates $(\rho/S,\,\theta/\pi,\,(\phi{+}\pi)/(2\pi))$\;
}

\caption{Per-scene volumetric data generation.}
\label{alg:dataset}
\end{algorithm}

\subsection{Houdini Dataset Construction}
\label{sec:supp_dataset}

We synthesize the training set in SideFX Houdini using its Pyro solver, covering three asset families: rising \emph{fire}, ambient \emph{smoke}, and bursting \emph{explosions}. Each scene is independently randomized in source geometry, solver parameters, wind perturbation, camera placement, and environment lighting, then rendered with the Karma path tracer to obtain RGB frames and a per-pixel velocity AOV. Algorithm.~\ref{alg:dataset} summarizes the per-scene pipeline; Tab.~\ref{tab:dataset_params} lists the randomization ranges that drive variation between sequences.

\mysuppparagraph{Source emitter.}
For fire and smoke we instantiate a noisy spherical emitter whose center is sampled from a Gaussian around the world origin. The noise displacement amplitude, pulse duration, fractal octaves and roughness are re-sampled per scene to diversify the initial density and temperature fields. Explosions reuse an authored \texttt{.hipnc} scene built around SideFX's pyro burst source and the Axiom GPU solver, with the \texttt{shapeoffset} seed randomized so each run starts from a different turbulent pattern.

\mysuppparagraph{Pyro and Axiom solvers.}
The Pyro solver runs at voxel size $\Delta x \!=\! 0.02$ for fire and $\Delta x \!=\! 0.01$ for smoke. Each scene resamples buoyancy lift, turbulence strength, disturbance strength, divergence scale, and dissipation; fire additionally resamples the blackbody intensity. Explosions are driven by the Axiom solver. All scenes simulate $T \!=\! 90$ frames at $24\,\text{fps}$.

\mysuppparagraph{Wind perturbation.}
With probability of $\frac{1}{2}$ we inject one rasterized wind volume into the solver. A box of random extent is scattered with points, each carrying a velocity vector $\mathbf{w} \!=\! s \cdot \hat{\mathbf{d}}$
where $s \!\sim\! \mathcal{U}(0, 0.8)$ and the azimuth of
$\hat{\mathbf{d}}$ is drawn uniformly on $[0, 2\pi)$; the field is
rasterized and connected as an additional
velocity source to the Pyro solver.

\mysuppparagraph{Camera and lighting.}
A single perspective camera is placed at
$(0,\,y,\,z)$ with $z \!\sim\! \mathcal{U}(6,10)$ for fire/smoke and
$z \!\sim\! \mathcal{U}(30,35)$ for explosions, oriented via a
look-at transformation to the emitter center. Lighting combines a random HDR
environment map drawn from a library with intensity
sampled per scene, and for smoke scenes, two opposing rim point lights
to better expose silhouettes.

\mysuppparagraph{Rendering and velocity AOV.}
We render with Karma renderer at $128$ path-traced samples. To export per-pixel velocity, we patch the Pyro shader's material network with two binded VOPs: the first reads the volume's \texttt{vel} attribute, the second exports it under the AOV name \texttt{velocity}. This produces a second image in the same EXR whose three channels are the velocity at that pixel.

\end{document}